\documentclass[runningheads]{llncs}

\usepackage{eccv}
\usepackage[colorlinks,urlcolor=eccvblue]{hyperref}

\usepackage{eccvabbrv}

\usepackage{graphicx}
\usepackage{booktabs}

\usepackage[accsupp]{axessibility}  

\usepackage{hyperref}

\usepackage{orcidlink}

\usepackage{fontawesome5} 
\usepackage{xcolor}       

\begin{document}

\title{MetaView: Monocular Novel View Synthesis with Scale-Aware Implicit Geometry Priors} 

\titlerunning{MetaView}

\author{Yufei Cai\inst{1}, 
Xuesong Niu\inst{2}$^{*}$,  
Hao Lu\inst{3},
Kun Gai\inst{2},
Kai Wu\inst{2}$^{\dagger}$, 
Guosheng Lin\inst{1}$^{\dagger}$
}

\authorrunning{Y.~Cai et al.}

\institute{Nanyang Technological University \and
Kolors Team, KlingAI Research \and
The Hong Kong University of Science and Technology (Guangzhou)
}

\maketitle
{\let\thefootnote\relax\footnotetext{{$^{*}$Project lead. $^{\dagger}$Corresponding author.}}}


\begin{center}
    \vspace{-0.6em} 
    \begin{tabular}{r l}
        \faGlobe\ \textbf{Project: } & \href{https://prototypenx.github.io/MetaView/}{\texttt{https://prototypenx.github.io/MetaView}} 
    \end{tabular}
\end{center}

\vspace{-1.8em}
\begin{abstract}
Current visual generation models are capable of producing high-quality content, yet they lack a coherent perception of the spatial structure. Existing generative novel view synthesis methods typically introduce explicit geometry priors, which enforce spatial consistency but inherently restrict generalization in large view changes. In contrast, recent interactive generative methods favor implicit scene modeling, offering greater flexibility at the cost of precise camera control and geometry consistency. In this paper, we propose \textbf{MetaView}, a diffusion-based monocular novel view synthesis framework that enables rendering under large view changes from a single image. Our key insight is to combine implicit geometry modeling with minimal yet essential explicit 3D cues: we incorporate implicit geometry priors from a feed-forward geometry perception network to regularize structure without imposing restrictive reconstruction pipelines, while leveraging metric depth to anchor the generation to a metric scale. This design allows MetaView to achieve both geometry consistency and precise controllability. Extensive experiments demonstrate that, under challenging monocular large viewpoint changes, MetaView significantly outperforms existing methods and exhibits superior generalization. Our code is publicly available at \href{https://github.com/KlingAIResearch/MetaView}{https://github.com/KlingAIResearch/MetaView}.

  \keywords{Visual Generation \and Novel View Synthesis \and Camera Control}
\end{abstract}
\vspace{-2.4em}








\begin{figure}[tb]
  \centering
  \includegraphics[width=1\linewidth]{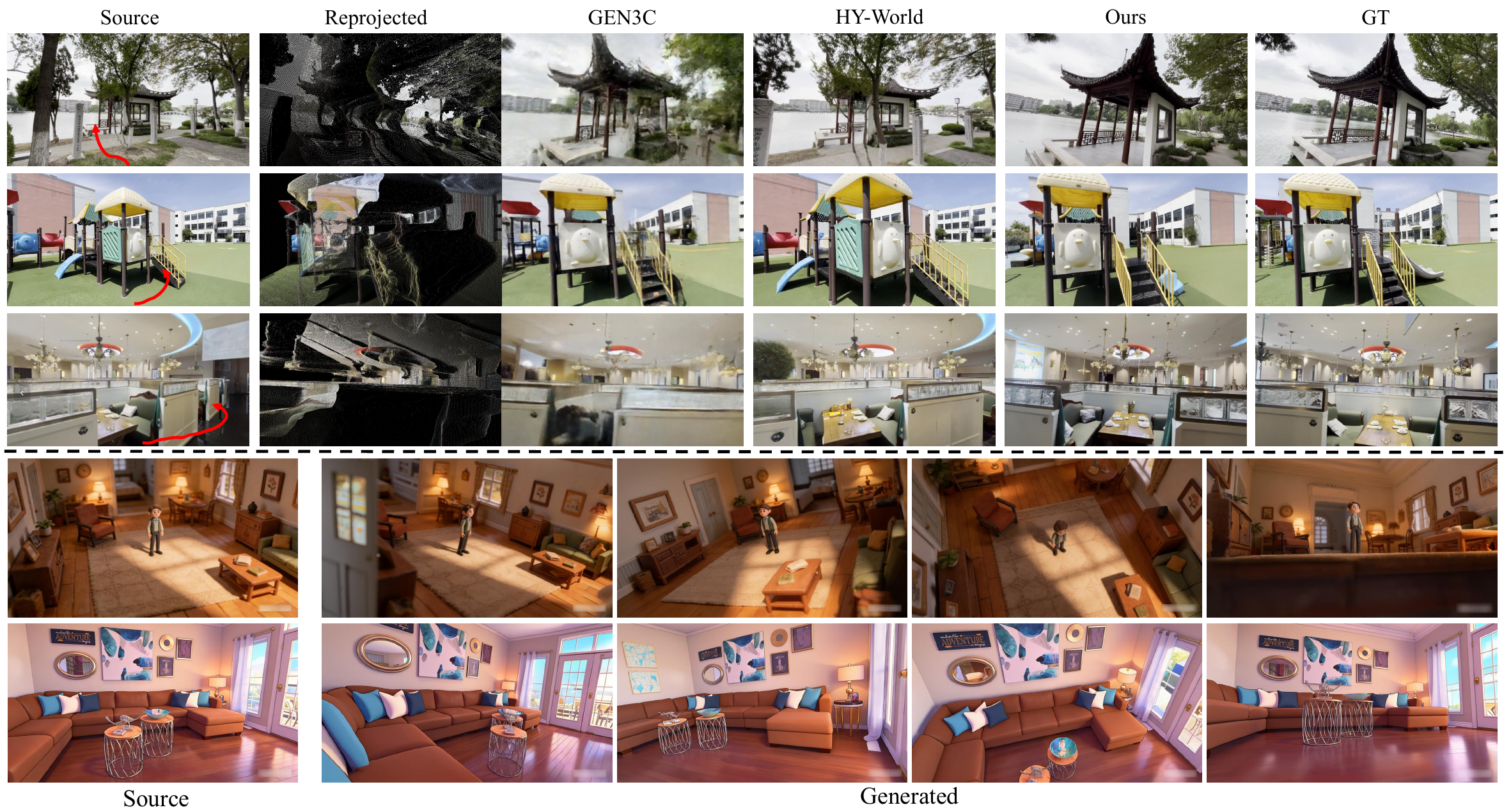}
  \caption{\textbf{Monocular novel view synthesis results.} \textbf{Upper}: Visual comparisons across different methods.  \textbf{Lower}: Generalization of our MetaView in diverse scenes.}
  \label{fig:teaser}
  \vspace{-1.2em}
\end{figure}

\section{Introduction}
\label{sec:intro}

Benefiting from the proliferation of large-scale visual data, feed-forward paradigms have emerged as an effective and scalable approach to 3D understanding and generation. In 3D perception, recent works\cite{dust3r,vggt,DA3} show that direct mapping from images to geometric representations offers superior efficiency and robustness compared to traditional optimization-based pipelines. This trend has naturally extended to visual generation, where diffusion-based\cite{DDPM,flowmatching} models reformulate scene reconstruction as a data-driven pixel prediction problem. A pivotal yet challenging task within this domain is monocular novel view synthesis (NVS), which aims to synthesize target views under arbitrary camera poses given a single source input as shown in Fig. \ref{fig:teaser}. While existing methods attempt to address this by either incorporating explicit 3D representations\cite{gen3c,viewcrafter} or learning spatial geometry in a purely implicit manner\cite{HY-world,matrixworld}, they often struggle to balance consistency with generalization capability under large viewpoint changes.

\begin{figure}[tb]
  \centering
  \includegraphics[width=1\linewidth]{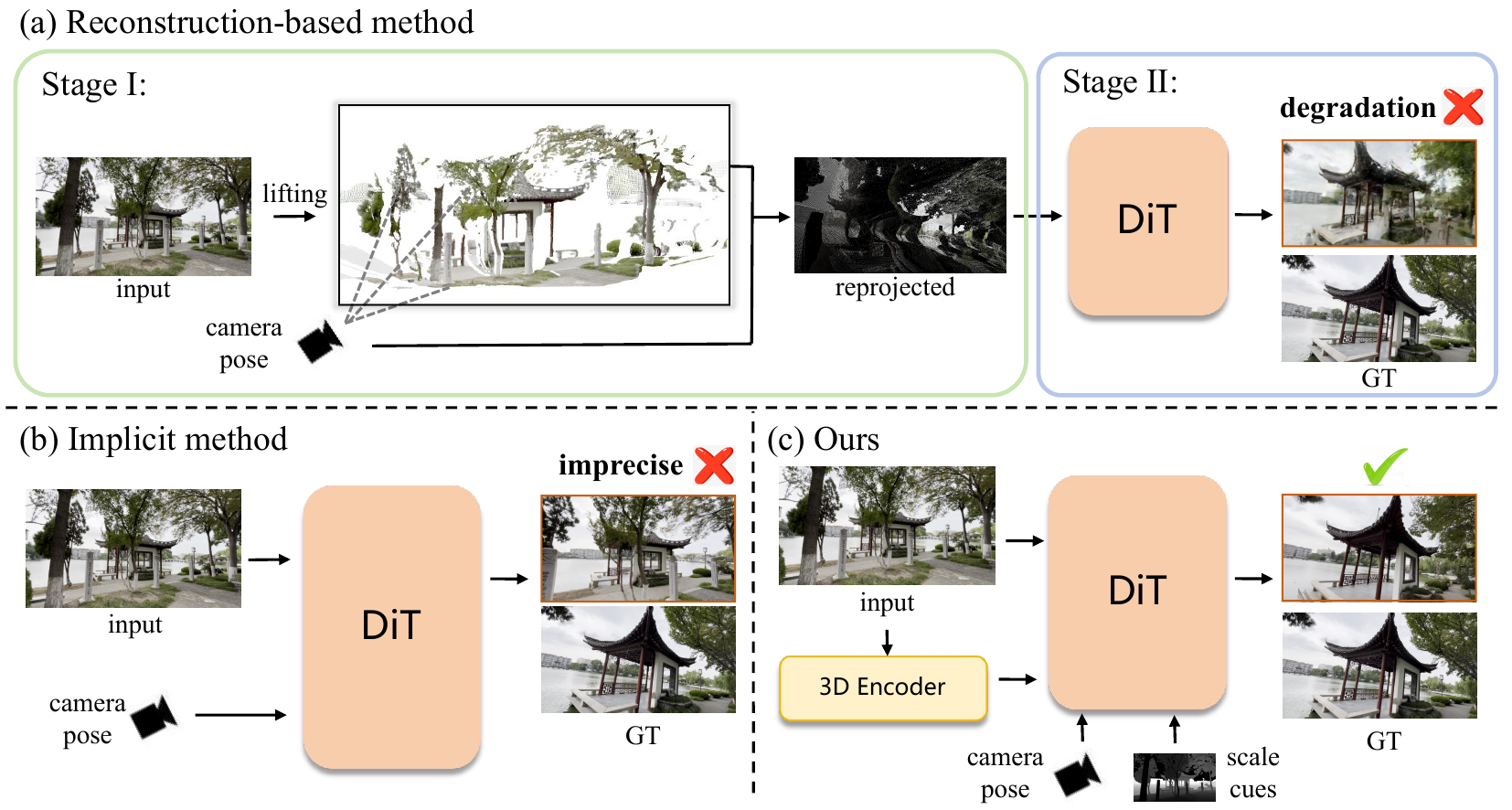}
  \caption{\textbf{Comparison of diffusion-based methods.} Heavy reliance on explicit 3D inductive biases leads to severe visual degradation, while fully implicit methods fail to achieve effective camera control. Our method effectively addresses both issues.}
  \label{fig:motivation}
  \vspace{-1.2em}
\end{figure}

As shown in Fig. \ref{fig:motivation}, existing diffusion-based NVS methods generally fall into two categories. One line of research adopts a two-stage pipeline\cite{uni3c,viewcrafter,gen3c,voyager,PE-field}: an explicit reconstruction module first establishes a sparse scene representation, which is then reprojected or warped to the target viewpoint to provide geometric conditions for a diffusion model. Such two-stage pipelines acquire explicit priors through reconstruction, after which the diffusion model mainly performs inpainting rather than spatial reasoning. While these explicit inductive biases ensure local consistency, they are inherently constrained by the quality of sparse reconstruction. Conversely, more recent works\cite{matrixworld,HY-world,lingbot-world} attempt to construct visual world models to synthesize novel views in a purely implicit manner by directly conditioning on camera poses. Despite the remarkable progress in interactive generation, their 3D perception (e.g., scene depth) remains ambiguous, leading to a scale drifting issue and spatial inconsistency during camera control.

We argue that enhancing spatial awareness in implicit generation models requires moving beyond this dichotomy. Reliance on explicit reconstruction constrains generalization, while learning 3D structure purely from pixel supervision is underdetermined. To this end, we present MetaView, a diffusion-based framework that achieves precise camera control and high-fidelity synthesis without the burden of explicit 3D reconstruction. Our key insight is to incorporate implicit geometry priors and minimal yet essential 3D inductive biases to regularize the generation process, endowing the model with improved spatial sensitivity and robust monocular novel view synthesis capacity.
Specifically, we extract hierarchical features from a feed-forward geometry perception network (e.g., DepthAnything3\cite{DA3}) as implicit priors to regularize generated scene structure with relative geometry relations. Following PRoPE\cite{prope}, we encode camera intrinsics and extrinsics into a modified Rotary Positional Encoding (RoPE)\cite{rope} to achieve camera control. Simultaneously, to resolve scale ambiguity, we also assign the metric scale cues of the input image into the modified RoPE by allocating an extra subspace for the $z$-axis. Built upon a pretrained multi-modal diffusion transformer (MM-DiT)\cite{sd3,qwen-image}, MetaView incorporates these geometric signals via non-invasive parallel attention layers, preserving the rich semantic knowledge of the base model while injecting structured 3D awareness.

To facilitate a faithful evaluation of NVS under extreme variations, we also introduce a new metric, Dense Matching Distance (DMD), which provides a more accurate assessment of spatial alignment than low-level metrics. We conduct extensive experiments on several datasets, including DL3DV\cite{dl3dv}, RealEstate10K\cite{realestate}, and Sekai-Real-Walking-HQ\cite{sekai}. 
Experimental results demonstrate that, MetaView significantly outperforms both reconstruction-based and purely implicit methods. In particular, our method exhibits superior generalization and more faithful geometric consistency under challenging large viewpoint changes. 

Our contributions can be summarized as follows:
\begin{itemize}
    \item We present MetaView, a robust monocular novel view synthesis method capable of handling large viewpoint changes without explicit reconstruction.
    \item We introduce implicit geometry priors to facilitate relative geometric consistency in complex scenes. Furthermore, we identify and mitigate the scale drifting issue inherent in implicit generative models by injecting scale cues.
    \item Extensive experiments demonstrate that MetaView consistently outperforms existing methods and exhibits strong generalization across diverse scenarios.
\end{itemize}

\section{Related Works}

\subsection{Text-to-Image Diffusion Models}
Diffusion model\cite{DDPM} is a parameterized neural network that learns to approximate complex high-dimensional distributions through an iterative denoising process, and has therefore become widely adopted in visual generation tasks. Latent diffusion model (LDM)\cite{LDM} performs denoising in a lower-dimensional latent space and introduces conditions through cross-attention mechanisms, achieving a favorable balance between generation efficiency and visual fidelity. Benefiting from large-scale data collection, pretrained text-to-image (T2I) diffusion models have demonstrated remarkable image generation capabilities. Initially, large-scale T2I models\cite{LDM,DALLE2,GLIDE,Imagen,sdxl} adopt U-Net\cite{unet} architectures as the denoising backbone. However, as model scales increased, Transformer-based DiT\cite{dit} emerged as a more scalable alternative, offering improved generalization and flexibility. Meanwhile, the flow-matching paradigm\cite{rectified_flow,flowmatching} gained traction due to its streamlined theoretical formulation and stable training dynamics, and has been widely adopted in modern pretrained models. Recent T2I models\cite{flux, qwen-image,zimage,sd3} typically employ a dual-stream Multi-Modal DiT (MM-DiT)\cite{sd3} architecture that separately models text and image tokens while enabling effective cross-modal interaction. This design substantially improves instruction-following capability and scalability. Our method builds upon Qwen-Image-Edit\cite{qwen-image}, a variant T2I model tailored for reference image editing. It augments the image stream by concatenating reference image tokens during forward propagation, allowing the model to perform generation conditioned on both text and visual inputs.

\subsection{Feed-forward Geometry Estimation}
Traditional geometry estimation systems\cite{sfm,msfm,mvsnet,xu2023iterative} decompose 3D reconstruction into handcrafted feature extraction and joint optimization pipelines. Despite the good performance in general scenes, they often suffer from limited robustness and scalability. The emergence of DUSt3R\cite{dust3r} shifts this paradigm by directly performing implicit modeling using a transformer, predicting camera poses and depth maps in a feed-forward manner. Subsequent works\cite{must3r,mv-dust3r,fast3r,monst3r} extend this framework to handle a larger number of inputs, enabling geometry estimation over broader scenes. VGGT\cite{vggt} and its variants\cite{pivggt,vggt-long,vggt-slam} further advance feed-forward geometry estimation by leveraging multi-task supervision and large-scale pretraining, substantially pushing the performance to a new level. More recently, DepthAnything3\cite{DA3} adopts a minimal yet effective single-transformer architecture, achieving higher efficiency and improved estimation accuracy in multiple downstream tasks. Therefore, we employ the SOTA method DepthAnything3 as our implicit geometry prior network.

\subsection{Novel View Synthesis}
Novel View Synthesis (NVS)\cite{nvs1,nvs2,nvs3} is a long-standing problem in computer vision and graphics. The emergence of Neural Radiance Fields (NeRF)\cite{nerf} marked a breakthrough in NVS. NeRF parameterizes 3D representations via implicit neural networks and synthesizes images via volume rendering. Subsequent research\cite{mip-nerf,pixelnerf,instant-ngp,nerf--,DAGO,mvsnerf} focused on improving rendering fidelity and accelerating optimization. 3D Gaussian Splatting (3DGS)\cite{3DGS} improves reconstruction efficiency and rendering quality by representing scenes as anisotropic Gaussian primitives. Later works\cite{mip-gs,pixelnerf,mvsplat,dreamgaussian} attempt to loosen requirements on the input views and incorporate neural decoders to enhance generalization.

More recently, diffusion-based generative models have emerged as an alternative framework for NVS\cite{zero123,GeNVS}. ZeroNVS\cite{zeronvs} leverages diffusion priors to perform 3D distillation. However, their generalization and geometric consistency remain limited. Subsquent works\cite{genwarp,PE-field} approache NVS through 2D coordinate warping, but lacks 3D awareness. Other works\cite{viewcrafter,gen3c,voyager,uni3c} introduce explicit 3D caches to guide generation via conditioned inpainting. While these methods improve camera control and cross-view consistency, their reliance on explicit reconstruction restricts generalization under sparse-view cases. A different line of work\cite{seva,anyview,lvsm,gcd} explores large-scale implicit models for scene generation. Recent works\cite{matrixworld,HY-world,lingbot-world} pursue interactive world generation through direct pixel-level regression. Although these methods employ camera pose as a control signal, their spatial geometry awareness remains ambiguous, resulting in imprecise viewpoint control. In this work, we provide the model with spatial awareness by injecting essential 3D cues and implicit geometric priors in a non-invasive manner, striking a balance between controllability and generalization.


\section{Method}
Given a single source image $X^{src}$ and a target camera pose $T$ defined relative to this image, our objective is to synthesize a novel view $X^{gen}$ of the scene $X^{src}$ under the viewpoint specified by $T$. In this section, we first present an overview of the DiT model used in our method (Sec.~\ref{sec:3.1}). Next, we introduce our approach for injecting explicit 3D cues via RoPE (Sec.~\ref{sec:3.2}) and explain the extraction and integration method of implicit geometric priors (Sec.~\ref{sec:3.3}). Finally, we describe the adaptation strategy for the pretrained generation model through parallel attention layers to incorporate these geometric signals (Sec.~\ref{sec:3.4}).

\subsection{Preliminary}
\label{sec:3.1}
We adopt pretrained Qwen-Image-Edit\cite{qwen-image} as our base model. It is a T2I model built upon the MM-DiT\cite{sd3} architecture and trained under the flow matching paradigm\cite{flowmatching,rectified_flow}, supporting both text and image conditions. The DiT network $v_\theta$ is composed of multiple stacked Multi-Modal Attention (MMA) blocks. Given a reference image and a text prompt, both modalities are first patchified into tokens $X=[X^{src}:X^{gen}] \in \mathbb{R}^{N \times\ d}$ and $C\in \mathbb{R}^{M \times\ d}$, respectively. These tokens are processed in a dual-stream manner, where modality-specific tokens are initially propagated through separate streams. In each MMA block, image and text tokens are projected into modality-specific $Q$, $K$, $V$, which are then concatenated along the sequence dimension and processed via self-attention mechanism\cite{attention}, enabling cross-modal interaction and information fusion:
\begin{equation}
\operatorname{Attention}([X; C]) = \operatorname{Softmax}\left( \frac{Q K^\top}{\sqrt{d}} \right) V,
\end{equation}

To preserve position relations, tokens are equipped with Rotary Positional Encoding (RoPE)\cite{rope}. For text tokens, 1D RoPE is applied to encode relative positions along the sequence. Both source image $X^{src}$ and generated image tokens $X^{gen}$ are assigned rotary transformations by $(k,i,j)$, where $k$ distinguishes different images, and $(i,j)$ denote the token’s coordinates on the 2D grid.


\begin{figure}[tb]
  \centering
  \includegraphics[width=1\linewidth]{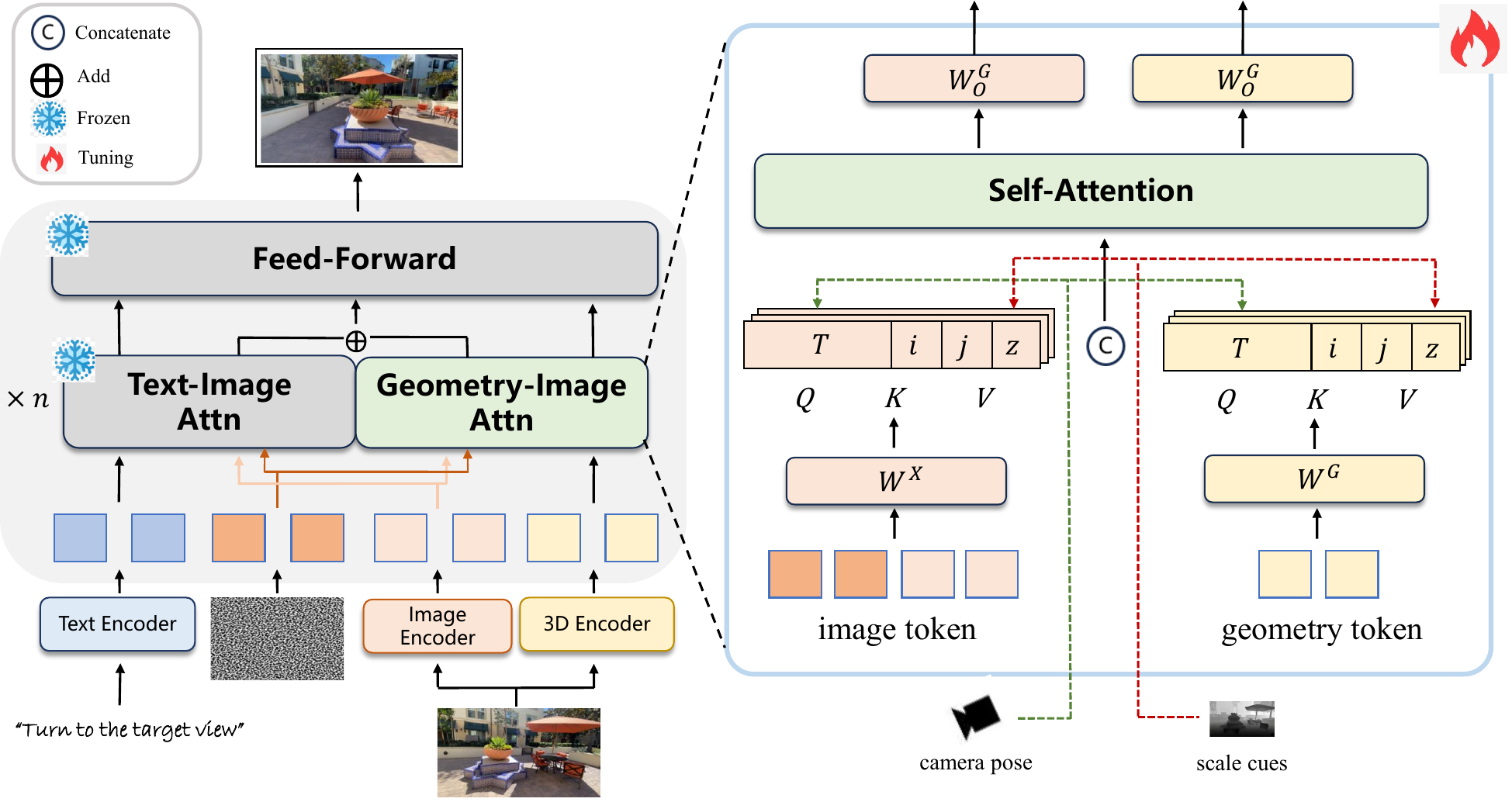}
  \caption{\textbf{Overview of MetaView.} \textbf{Left}: Our method is built upon the MM-DiT architecture, which integrates the implicit geometry priors by adding a split stream in the DiT. \textbf{Right}: We fuse geometry priors through self-attention mechanism, while injecting scale cues via a modified spatial-aware RoPE to address the scale drifting issue.}
  \label{fig:pipeline}
  \vspace{-1.2em}
\end{figure}



\subsection{3D Cues Injection}
\label{sec:3.2}
The input condition is defined as the set $\{X^{src},K,T\}$, where $X^{src} \in \mathbb{R}^{H \times W \times 3}$ denotes the input image, $K \in \mathbb{R}^{3 \times 3}$ is the corresponding camera intrinsic matrix, and 
$T=(R,t) \in SE(3)$ represents the relative extrinsic of the target viewpoint. The 6-DoF parameterization of $T$ encodes the camera position and orientation in the world coordinate. Given a 3D point $X_{world} \in \mathbb{R}^{4}$ in homogeneous world coordinates, its projection onto the image frustum can be formulated as:

\begin{equation}
\tilde{P} = 
\begin{bmatrix}
K & \mathbf{0}^{3 \times 1}
\end{bmatrix} 
T, \quad \tilde{X} = \tilde{P} X_\text{world} \in \mathbb{R}^3
\end{equation}
where $\tilde{X}$ is the projected homogeneous image coordinate.
Accurate projection via this formulation requires that $X_{world}$ and $T$ share a consistent metric scale. If the underlying scale of $X_{world}$ is misaligned with $T$, projection becomes ambiguous, leading to the \textbf{scale drifting} issue commonly observed in SLAM systems.

Reconstruction-based methods\cite{gen3c,voyager,viewcrafter,PE-field} typically estimate camera poses and depth jointly, ensuring that geometric quantities are aligned within a consistent metric scale. This alignment guarantees geometric consistency during view transformation. In contrast, diffusion-based implicit methods\cite{HY-world,lingbot-world} lack an explicit coupling between the provided camera pose and the scene scale internally captured by the model. Since generation models do not explicitly reconstruct 3D structure, the implicit geometry is scale-ambiguous, often resulting in unstable viewpoint control under pose variation. 

To mitigate this issue, we introduce the depth $z$ with metric scale and incorporate aligned camera poses during training to eliminate scale ambiguity. Specifically, to enable camera control, we follow PRoPE\cite{prope} to encode the intrinsic and extrinsic $K$ and $T$ into RoPE, allowing the model to capture frustum-aware geometric relationships. We adopt the off-the-shelf DepthAnything3-Metric\cite{DA3} to estimate metric depth $z$ from the input image, explicitly anchoring the generation process to consistent spatial scale cues. In implementation, we allocate an extra subspace within the RoPE to represent the $z$-axis component and assign the extracted $z$ to this dimension. The overall process can be formally expressed as follows:

\begin{equation}
\begin{aligned}
D_t^{\text{RoPE}} &= 
\begin{bmatrix}
D_t^{\text{Proj}} & \mathbf{0} \\
\mathbf{0} & D_t^{\text{Grid}}
\end{bmatrix}; \quad
D_t^{\text{Proj}} &= \mathbf{I}_{d/8} \otimes \tilde{P} \in \mathbb{R}^{\frac{d}{2} \times \frac{d}{2}},
\end{aligned}
\end{equation}

\begin{equation}
\begin{aligned}
D_t^{\text{Grid}} &= 
\begin{bmatrix}
\text{RoPE}_{d/6}(i_t) & \mathbf{0} & \mathbf{0} \\
\mathbf{0} & \text{RoPE}_{d/6}(j_t) & \mathbf{0} \\
\mathbf{0} & \mathbf{0} & \text{RoPE}_{d/6}(z_t)
\end{bmatrix}
\in \mathbb{R}^{\frac{d}{2} \times \frac{d}{2}}.
\end{aligned}
\end{equation}
where $\otimes$ denotes Kronecker product and $\mathbf{I}$ is identity matrix. $\text{RoPE}_{d/6}(\cdot)$ constructs rotary embeddings in $\frac{d}{6}$ dimensions for $(i,j,z)$ coordinates of token $t$. We follow GTA's\cite{GTA} formulation to transform $Q$, $K$ and $V$ in the attention:
\begin{equation}
\mathbf{D}^{\text{RoPE}} \odot \operatorname{Attn}\big( (\mathbf{D}^{\text{RoPE}})^\top \odot Q, (\mathbf{D}^{\text{RoPE}})^{-1} \odot K, (\mathbf{D}^{\text{RoPE}})^{-1} \odot V \big)
\end{equation}
where $\odot$ denotes matrix product.

Noting that we do not perform handcrafted reconstruction operations such as depth lifting, thereby reducing the extra explicit 3D inductive biases. Unlike methods such as PE-Field\cite{PE-field}, we do not rely on fine-grained depth to model relative geometry relationships. Instead, we downsample $z$ to $z'$ to match the spatial resolution of latent tokens in the attention layers and inject sparse cues $z'$ through PE. This design provides an explicit measure of spatial scale, aligning the model’s internal spatial perception with the camera geometry via RoPE.

%
\subsection{Implicit Relative Geometry Priors}
\label{sec:3.3}
Relative geometric relationships are crucial for novel view synthesis. Although diffusion models can implicitly capture certain relational geometry through pixel-level prediction, object distortions and pose deviations frequently arise in dense and structurally complex scenes. To mitigate these issues while avoiding excessive explicit 3D inductive biases that may hinder generalization, we incorporate fine-grained implicit relative geometry priors to guide the generation process. Rather than explicit reconstruction, our goal is to provide relative geometry cues that regularize spatial reasoning within the implicit backbone.

Feed-forward 3D perception models\cite{dust3r,vggt,DA3} trained with multi-task objectives naturally capture rich geometry of scene structure. In addition, the transformer-based architecture produce hierarchical intermediate features that encode geometry in a tokenized form, making it suitable for integration into DiT-based models. 

Motivated by this property, we leverage intermediate features from the feed-forward geometry model DepthAnything3\cite{DA3} as implicit geometry priors. 
Specifically, given the source image $X^{src}$, we extract intermediate features from selected layers that are originally fed into the decoder head for 3D prediction. These multi-level features are concatenated along the channel and projected to match the dimension $d$ with image tokens via a linear projection layer $W_G$:
\begin{equation}
G=W_G([\{f^{\phi}\}_{\phi \in decode}]) \in \mathbb{R}^{L \times d}
\end{equation}
Thus the geometry-awared $G$ are treated as geometry tokens, which are projected to $Q$, $K$, $V$ and then concatenated with image tokens $X=[X^{src};X^{gen}]$ in self-attention:
\begin{equation}
\operatorname{Attention}([X; G]) = \operatorname{Softmax}\left( \frac{Q K^\top}{\sqrt{d}} \right) V,
\end{equation}

To ensure consistency, tokens are all encoded under the same RoPE formulation introduced in Sec.~\ref{sec:3.2}. Similar to source image tokens $X^{src}$, $G$ is associated with the camera pose shifted to the origin, denoted as $T^o=\mathbf{I}$, and is also assigned with the downsampled sparse depth $z'$. In contrast, the generated image tokens $X^{gen}$ corresponding to the target view are bound to the relative target camera pose $T$, while their $z$-axis components are set to $0$ to avoid directly constraining the generated scene:
\begin{equation}
\text{PE}(X^{src})=\text{PE}(G)=\mathbf{D}^{\text{RoPE}}(T^o,z');\quad
\text{PE}(X^{gen})=\mathbf{D}^{\text{RoPE}}(T,\mathbf{0})
\end{equation}
Under this formulation, $G$ inherently encodes relative geometric priors and are additionally anchored with absolute scale through $z'$. The generated image tokens $X^{gen}$, conditioned on the relative transformation $T$, aggregating geometry-aware guidance by attending to tokens defined under $T^o$. This interaction enables geometry-aware generation while preserving the implicit nature of the overall framework.


\subsection{Geometry-Guided Parallel Attention Adaptation}
\label{sec:3.4}
Our method is built upon a pretrained MM-DiT\cite{sd3,qwen-image} backbone. To maximally preserve its rich semantic knowledge, we freeze all pretrained parameters, thereby preventing catastrophic forgetting. Instead, we introduce parallel attention layers to inject both implicit geometry priors and scale cues.

Specifically, we construct an additional geometry stream for the introduced geometry tokens $G$, which feed forward alongside the original image and text streams across DiT blocks. To maintain compatibility with the original MM-DiT architecture, we do not alter the existing image–text attention. Instead, within each block, we introduce a parallel image-geometry attention layer that models interactions between $X$ and $G$. The resulting features are added to fuse with the original attention outputs, enabling structured 3D cue integration:
\begin{equation}
Out = \operatorname{Attention}([X; C]) +\operatorname{Attention}([X; G])
\end{equation}

For the image–geometry attention layer, we introduce a dedicated set of projection parameters $W_Q^X, W_K^X, W_V^X, W_O^X$ and $W_Q^G, W_K^G, W_V^G, W_O^G$ which project the image tokens $X$ and geometry tokens $G$ into their respective query, key, value embeddings and output projection. To ensure parameter efficiency, $G$ and $X$ share the same fixed feed-forward network (FFN) within each transformer block. In addition, a linear projection layer 
$W_G$ is applied before feeding into the network to map geometry tokens to the same channel dimension $d$ as image tokens. During training, only these newly introduced parameters are optimized, while all pretrained backbone weights remain frozen. The model is trained under the flow matching\cite{flowmatching} objective:

\begin{equation}
\mathcal{L}_{\text{FM}}(\theta) =
\mathbb{E}_{t,x_0,x_1,C,G}
\Bigl\|
v_\theta\bigl(x_t;C;G)
- (x_1-x_0)
\Bigr\|^2_2
\label{eqn:fm_loss}
\end{equation}

\begin{figure}[tb]
  \centering
  \includegraphics[width=1\linewidth]{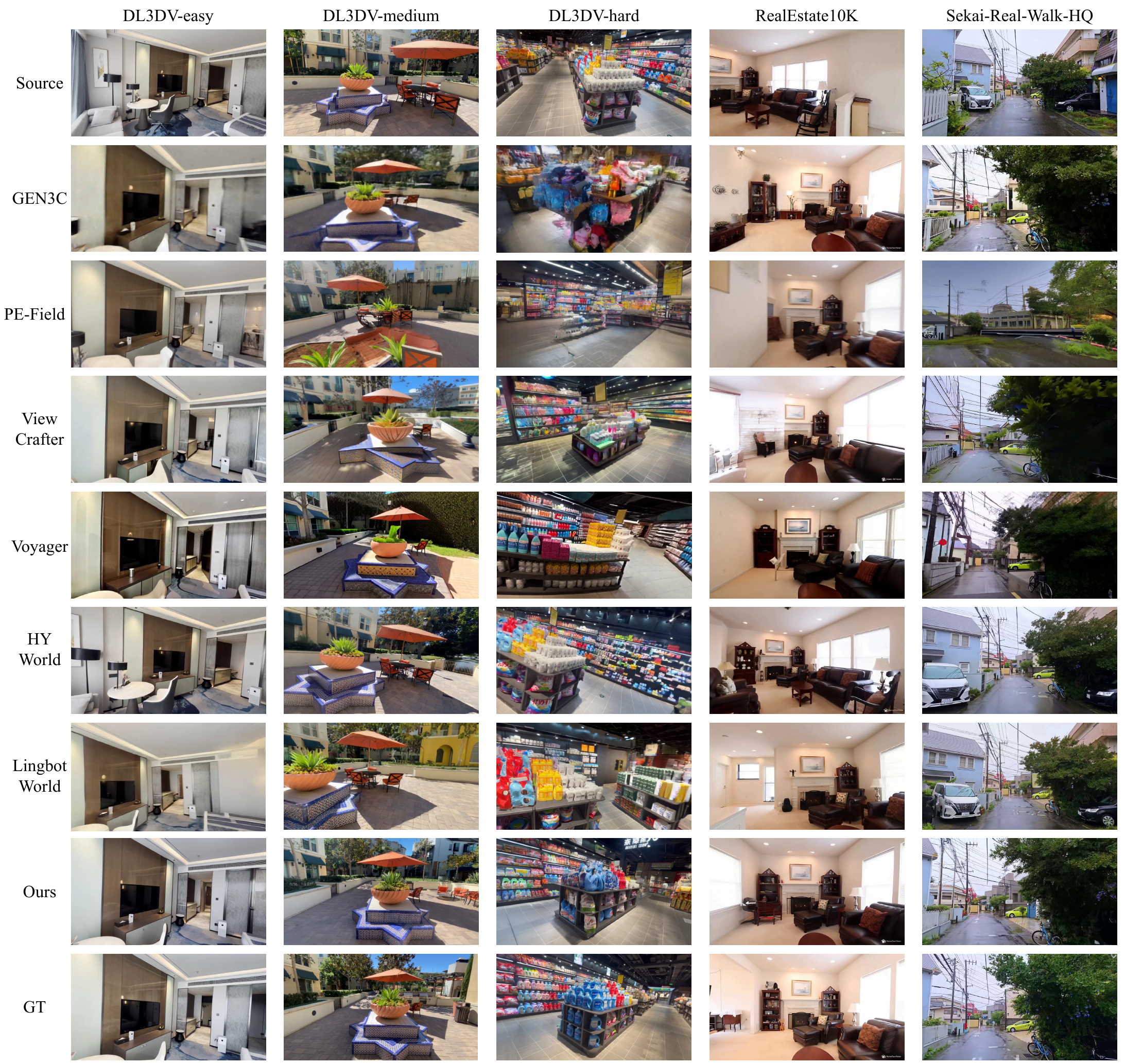}
  \caption{\textbf{Visual comparisons.} Our method enables more precise camera control and significantly outperforms baseline methods under large viewpoint changes.}
  \label{fig:qualitative}
  \vspace{-1.8em}
\end{figure}

\section{Experiments}

\subsection{Settings}

\noindent
\textbf{Datasets}. 
We conduct experiments on DL3DV\cite{dl3dv}, RealEstate10K\cite{realestate}, and Sekai-Real-Walking–HQ\cite{sekai} datasets. We collect only the RGB data from these datasets and estimate camera poses using VIPE\cite{vipe}. During pose estimation, VIPE adopts DepthAnything3–Metric\cite{DA3} as scale priors to align poses with metric depth. We further filter out samples with erroneous estimates and low-quality scenes. 
We evaluate the methods on the three aforementioned datasets, ensuring that test scenes do not overlap with the training set.
To assess performance under varying degrees of viewpoint change, we further partition the DL3DV test set into three difficulty levels—easy, medium, and hard—based on the view overlap ratio between the source and target images. Specifically, the easy subset contains pairs with more than 80\% overlap, the medium subset includes pairs with 50\%–80\% overlap, and the hard subset consists of pairs with 30\%–50\% overlap. For the RealEstate and Sekai test sets, the view overlap ratio is controlled within the range of 30\%–80\% to ensure diverse yet meaningful viewpoint variations. Details of data preparation are provided in the \textit{Suppl}.

\noindent
\textbf{Implementation details.}
We adopt Qwen-Image-Edit\cite{qwen-image} as the base model and use DepthAnything3-Giant\cite{DA3} to extract geometry priors. All experiments are conducted 
with a total batch size of 32. We employ the AdamW\cite{adamw} optimizer with a learning rate of 5e-5. The text prompts are fixed to "Turn to the target view".
For inference, we use 40 sampling steps with a classifier-free guidance\cite{cfg} scale of 4.

\noindent
\textbf{Baselines}. We compare our method with both reconstruction-based method: ViewCrafter\cite{viewcrafter}, Gen3C\cite{gen3c}, PE-Field\cite{PE-field}, Voyager\cite{voyager}, and implicit method: HY-World-1.5\cite{HY-world}, Lingbot-World\cite{lingbot-world}. All baselines are reproduced using the official open-source implementations with default configurations.

\noindent
\textbf{Metric}
We evaluate the methods with low-level metrics PSNR, SSIM\cite{ssim} and perceptual metric LPIPS\cite{lpips}. However, the above NVS metrics primarily evaluate performance under narrow viewpoint changes, where novel views are generated via interpolation. When the viewpoint variation becomes large and requires extrapolative generalization, these metrics fail to adequately reflect the capability.
Inspired by WorldScore\cite{worldscore}, we propose a new metric for NVS under arbitrary viewpoint variations, termed Dense Matching Distance (DMD). It measures the average optical flow displacement between matched points across two views. Specifically, we employ UFM\cite{ufm} to compute the co-visible regions and establish dense correspondences between the two views. Let $\text{UFM}(A; B)$ denote the set of points in $A$ matched to $B$ and their corresponding optical flow displacements $dist(p)$ respect to $B$ computed via UFM, we respectively obtain the matched point sets for the source and generated images relative to the ground truth $P_{src}=\text{UFM}(X^{src}; X^{GT})$ and $P_{gen}=  \text{UFM}(X^{gen}; X^{GT})$.
For each $p$ in $P_{src}$, if it also exists in $P_{gen}$, its optical flow distance is contributed to the metric; otherwise, a maximum distance penalty $\sigma$ is imposed:


\begin{equation}
\text{DMD}(X^{src};X^{gen};X^{GT}) = \frac{1}{|P_{\text{src}}|} \sum_{p \in P_{src}}
\begin{cases}
\lVert \text{dist}(p) \rVert_2^2, & p \in P_{gen} \\
\sigma, & p \notin P_{gen}
\end{cases}
\end{equation}
The reported metric values are normalized to the range of $0\sim100$. More analysis about the metric is provided in the \textit{Suppl.}

\begin{table}[tb]
    \caption{\textbf{Quantitative comparisons across difficulty on DL3DV.}}
    \label{tab:quantitative_DL3DV}
    \centering
    \resizebox{\linewidth}{!}{
    \begin{tabular}{l|cccc|cccc|cccc}
    \toprule
    Method & \multicolumn{4}{c|}{DL3DV-Easy~\cite{dl3dv}} & \multicolumn{4}{c|}{DL3DV-Medium~\cite{dl3dv}} & \multicolumn{4}{c}{DL3DV-Hard~\cite{dl3dv}} \\
    \cmidrule(lr){2-5} \cmidrule(lr){6-9} \cmidrule(lr){10-13}
    & PSNR$\uparrow$ & SSIM$\uparrow$ & LPIPS$\downarrow$ & DMD$\downarrow$ & PSNR$\uparrow$ & SSIM$\uparrow$ & LPIPS$\downarrow$ & DMD$\downarrow$ & PSNR$\uparrow$ & SSIM$\uparrow$ & LPIPS$\downarrow$ & DMD$\downarrow$ \\
    \midrule
    ViewCrafter~\cite{viewcrafter} & 15.45 & 0.4454 & 0.2037 & 8.84 & 13.19 & 0.3679 & 0.2733 & 24.35 & 11.68 & 0.3449 & 0.3223 & 48.83\\
    Gen3C~\cite{gen3c} & 15.32 & 0.4339 & 0.1964 & 6.52 & 14.14 & 0.4092 & 0.2556 & 12.43 & 11.87 & 0.3424 & 0.3089 & 41.07\\
    Voyager~\cite{voyager}& 15.19 & 0.4171 & 0.2530 & 12.21 & 13.45 & 0.3610 & 0.2886 & 23.89 & 11.24 & 0.3302 & 0.3462 & 54.26\\
    PE-Field~\cite{PE-field} & 15.01 & 0.3964 & 0.2129 & 8.91 & 13.18 & 0.3435 & 0.2749 & 15.41 & 11.97 & 0.3422 & 0.3278 & 41.46\\
    HY-World-1.5~\cite{HY-world} & 12.27 & 0.3323 & 0.2906 & 14.58 & 11.34 & 0.2948 & 0.3304 & 18.50 & 10.71 & 0.3043 & 0.3520 & 34.45\\
    Lingbot-World~\cite{lingbot-world} & 11.82 & 0.3191 & 0.3073 & 29.20 & 11.99 & 0.2912 & 0.3141 & 21.37 & 11.13 & 0.3319 & 0.3447 & 55.39\\
    \midrule
    Ours & \textbf{17.94} & \textbf{0.5542} & \textbf{0.1397} & \textbf{2.56} & \textbf{15.05} & \textbf{0.4225} & \textbf{0.2132} & \textbf{7.57} & \textbf{12.54} & \textbf{0.3802} & \textbf{0.2881} & \textbf{20.74}\\
    \bottomrule
    \end{tabular}
    }
\end{table}

\begin{table}[tb]
    \caption{\textbf{Quantitative comparisons on different datasets.}
    }
    \label{tab:quantitative}
    \centering
    \resizebox{\linewidth}{!}{
    \begin{tabular}{l|cccc|cccc|cccc}
    \toprule
    Method & \multicolumn{4}{c|}{DL3DV~\cite{dl3dv}} & \multicolumn{4}{c|}{RealEstate10K~\cite{realestate}} & \multicolumn{4}{c}{Sekai-Real-Walk-HQ~\cite{sekai}} \\
    \cmidrule(lr){2-5} \cmidrule(lr){6-9} \cmidrule(lr){10-13}
    & PSNR$\uparrow$ & SSIM$\uparrow$ & LPIPS$\downarrow$ & DMD$\downarrow$ & PSNR$\uparrow$ & SSIM$\uparrow$ & LPIPS$\downarrow$ & DMD$\downarrow$ & PSNR$\uparrow$ & SSIM$\uparrow$ & LPIPS$\downarrow$ & DMD$\downarrow$ \\
    \midrule
    ViewCrafter~\cite{viewcrafter} & 13.44 & 0.3861 & 0.2664 & 27.34 & 13.16 & 0.4994 & 0.2597 & 10.92 & 15.73 & 0.4388 & 0.3068 & 17.93 \\
    Gen3C~\cite{gen3c} & 13.78 & 0.3952 & 0.2536 & 20.01 & 14.07 & 0.5162 & 0.2306 & 9.01 & 16.56 & 0.4659 & 0.2568 & 8.72\\
    Voyager~\cite{voyager}& 13.29 & 0.3694 & 0.2959 & 30.12 & 14.55 & 0.5170 & 0.2519 & 12.95 & 16.24 & 0.4402 & 0.3205 & 19.07\\
    PE-Field~\cite{PE-field} & 13.39 & 0.3607 & 0.2719 & 21.93 & 14.51 & 0.5191 & 0.2354 & 14.81 & 15.74 & 0.4418 & 0.2900 & 19.35 \\
    HY-World-1.5~\cite{HY-world} & 11.44 & 0.3105 & 0.3243 & 22.51 & 12.34 & 0.4466 & 0.3040 & 13.91 & 14.62 & 0.4088 & 0.3240 & 15.04\\
    Lingbot-World~\cite{lingbot-world} & 11.65 & 0.3141 & 0.3220 & 35.32 & 12.26 & 0.4554 & 0.3026 & 24.51 & 14.52 & 0.3973 & 0.3258 & 18.71\\
    \midrule
    Ours & \textbf{15.18} & \textbf{0.4456} & \textbf{0.2137} & \textbf{10.29} & \textbf{15.27} & \textbf{0.5705} & \textbf{0.1980} & \textbf{6.44} & \textbf{17.72} & \textbf{0.5047} & \textbf{0.2333} & \textbf{6.69}\\
    \bottomrule
    \end{tabular}
    }
\end{table}

\subsection{Qualitative Evaluation}
As shown in Fig.~\ref{fig:qualitative}, MetaView demonstrates effective spatial awareness, producing coherent results under arbitrary viewpoint changes (Cols. 2, 3), while achieving precise control over camera pose and scene scale (Cols. 1, 5). In contrast, fully implicit methods\cite{HY-world,lingbot-world} suffer from scale drifting and inaccurate viewpoint control due to scale ambiguity. Reconstruction-based methods\cite{gen3c,viewcrafter,voyager,PE-field} perform well under narrow viewpoint changes but exhibit blurriness and distortions (Col. 1) due to sparse reconstruction limits. Under large viewpoint variations, these explicit geometric conditions fail, leading to severe synthesis degradation. Therefore, MetaView outperforms existing methods, demonstrating superior accuracy and generalization. Please refer to the \textit{Suppl} for more results.
\begin{figure}[tb]
  \centering
  \includegraphics[width=1\linewidth]{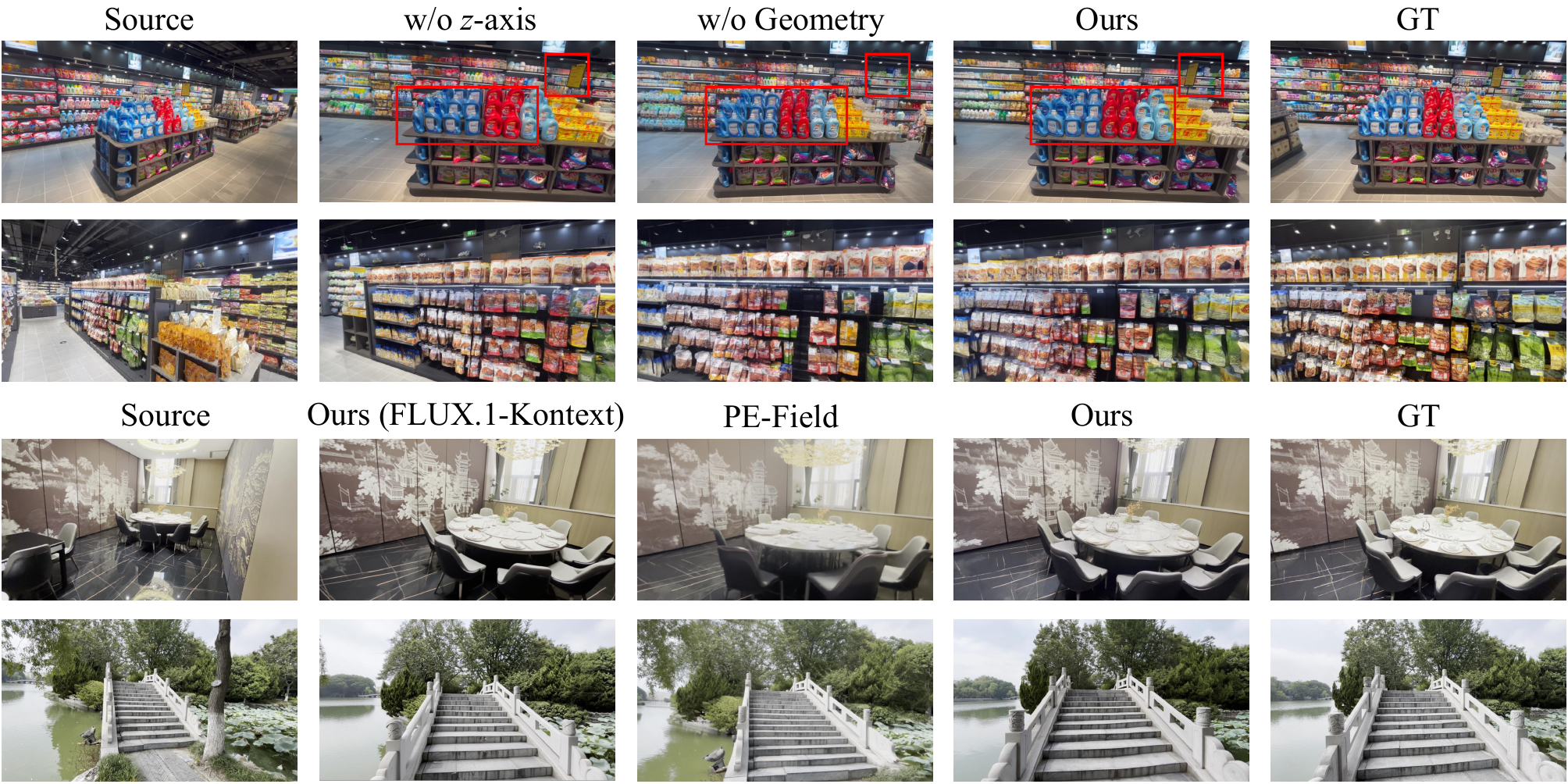}
  \caption{\textbf{Visualization of the ablation.} Scale cues provide the model with accurate spatial awareness, while geometry priors facilitate fine-grained relative consistency.}
  \label{fig:ablation}
    \vspace{-1.2em}
\end{figure}

\subsection{Quantitative Evaluation}
Tab.~\ref{tab:quantitative_DL3DV} presents the quantitative results on the DL3DV dataset across different difficulties. Reconstruction-based methods\cite{gen3c, viewcrafter, voyager,PE-field} perform well on the easy subset but degrade significantly as difficulty increases. Implicit methods\cite{HY-world,lingbot-world} perform poorly across all subsets due to imprecise viewpoint control, showing limited sensitivity to large view variations. MetaView achieves the best performance across all difficulties. Notably, on the hard subset, low-level metrics (PSNR, SSIM) become less indicative of perceptual differences, whereas our proposed DMD metric more clearly highlights the performance gap. Furthermore, MetaView consistently yields the best overall results on the RealEstate and Sekai datasets (Tab.~\ref{tab:quantitative}), demonstrating strong robustness and generalization.


\subsection{Ablation Study}

We conduct ablations on the geometry tokens, the z-axis in RoPE and the pretrained backbone. To ensure efficiency, we conduct the following experiments on the DL3DV-medium test set. Fig. \ref{fig:ablation} presents qualitative results after ablating each component, while Tab. \ref{tab:ablation} reports the corresponding quantitative impact.


\noindent
\textbf{Effect of Geometry Token.}
We first ablate the implicit geometry priors by removing the geometry tokens. As shown in Fig. \ref{fig:ablation} (Col. 3), although the overall viewpoint orientation remains approximately correct, the model struggles to preserve complex relative relationships within the scene. For example, the signboard disappears and the fine-grained contours of objects have noticeable discrepancies. In the second row, the relative arrangement of objects deviates from the ground truth. 
Quantitatively, removing the geometry tokens results in performance degradation across all evaluation metrics. This decline aligns with our qualitative observations, as the absence of geometry priors weakens the model’s ability to reason about fine-grained relative structure.

\begin{table}[tb]
    \caption{\textbf{Ablation study.} We conduct the training on DL3DV and testing on the DL3DV-medium. The effects of each component are reflected in the evaluation metrics.}
    \label{tab:ablation}
    \centering
    \resizebox{0.65\linewidth}{!}{
    \begin{tabular}{l|cccc}
    \toprule
     & PSNR$\uparrow$ & SSIM$\uparrow$ & LPIPS$\downarrow$ & DMD$\downarrow$ \\
    \midrule
    Ours w/o geometry & 13.53 & 0.3538 & 0.2530 & 11.61 \\
    Ours w/o $z$-axis & 12.49 & 0.3204 & 0.2900 & 14.45 \\
    Ours w/ FLUX.1-Kontext & 13.90 & 0.3594 & 0.2571 & 10.22\\ 
    PE-Field (FLUX.1-Kontext) & 13.18 & 0.3435 & 0.2749 & 15.41 \\
    Ours & \textbf{14.21} & \textbf{0.3765} & \textbf{0.2353} & \textbf{9.33} 
    \\
    \bottomrule
    \end{tabular}
    }
\end{table} 

\noindent
\textbf{Effect of $z$-axis.}
We ablate the $z$-axis in RoPE by conditioning PE solely on pose and coordinates $(T, i, j)$. As shown in Fig.~\ref{fig:ablation}, removing the $z$-axis leads to a degradation in camera control accuracy. While the rotation remains correct, the viewpoint position exhibits a translation offset from the ground truth, reflecting the scale drifting issue. Incorporating the $z$-axis anchors scene scale and restores accurate control. Quantitatively (Tab.~\ref{tab:ablation}), removing it consistently degrades all metrics, particularly the low-level metrics PSNR and SSIM. This confirms that $z$-axis scale cues are essential for mitigating scale drifting issue and enabling scale-aware spatial reasoning.

\noindent
\textbf{Effect of Backbone.} We adapt MetaView to the FLUX.1-Kontext backbone with the results summarized in Tab.~\ref{tab:ablation} and Fig.~\ref{fig:ablation}. Using the same base model, MetaView outperforms PE-Field. While there is a moderate performance drop compared to our Qwen-Image-Edit variant, this degradation is primarily attributed to the inherent capability gap between the backbones. This confirms the effectiveness of our design and demonstrates the robustness of our framework in cross-model adaptation.

\subsection{Analysis of DMD Metric}
\begin{figure}[tb]
  \centering
  \includegraphics[width=1\linewidth]{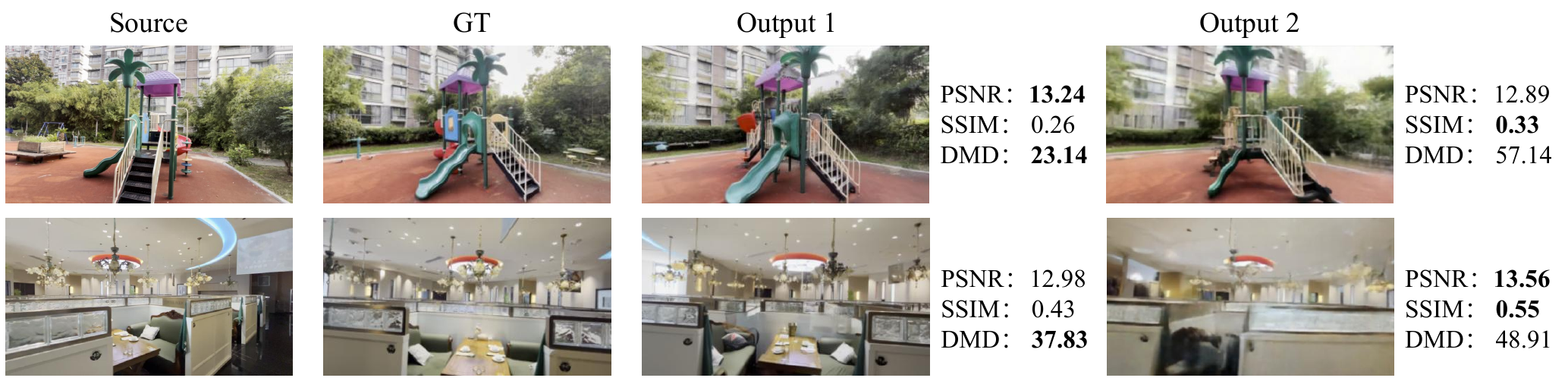}
  \caption{\textbf{Comparison of metric reliability.} Low-frequency structural alignment dominates PSNR and SSIM, while our proposed DMD aligns with human preference.}
  \label{fig:metric}
  \vspace{-1.2em}
\end{figure}

Low-level metrics fail in large-view monocular NVS, as extrapolated regions require plausible completion rather than strict pixel alignment. They are dominated by low-frequency alignment (Fig. \ref{fig:metric}), assigning higher scores to blurry artifacts (Output 2) and contradicting human perception.
In contrast, our DMD aligns with human preference. In both cases, DMD correctly favors the visually superior Output 1. The numerical margin further demonstrates its discriminative ability in evaluating extrapolative synthesis.

\begin{figure}[tb]
  \centering
  \includegraphics[width=1\linewidth]{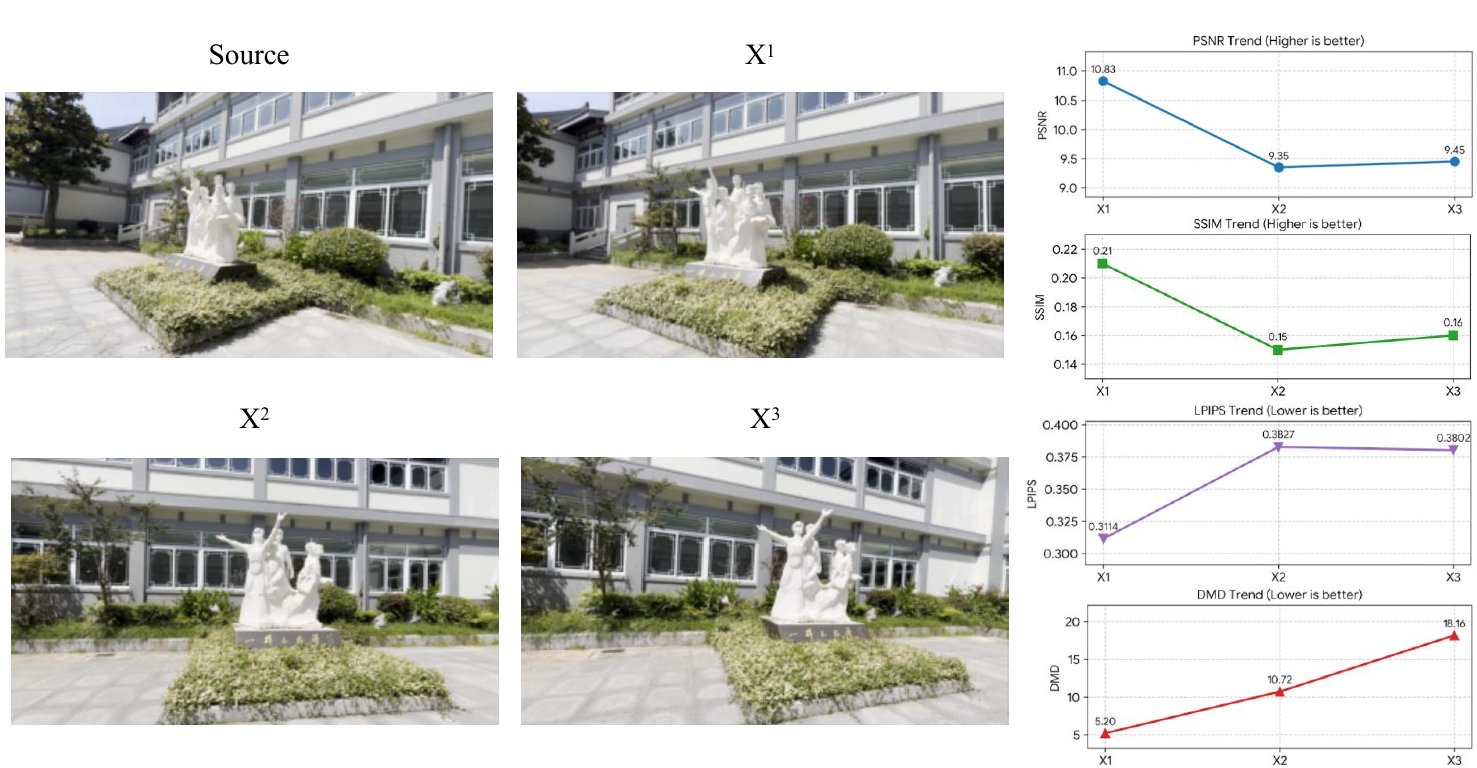}
  \caption{\textbf{Trends of Different Metrics.} While traditional metrics (PSNR, SSIM, and LPIPS) fail to show a monotonic trend as the viewpoint changes, our proposed DMD exhibits monotonicity, accurately quantifying the discrepancy between views.}
  \label{fig:metric2}
  \vspace{-1.2em}
\end{figure}

Furthermore, DMD can numerically reflect the magnitude of viewpoint variations. As illustrated in Fig. \ref{fig:metric2}, given the source image $X^{src}$, let $X^i$ denote a sequence of views with progressively increasing viewpoint changes. We compute the PSNR, SSIM, and LPIPS between each $X^i$ and $X^{src}$, along with our proposed metric $\text{DMD}(X^{i};X^{i};X^{src})$. It can be observed that as the viewpoint variation increases, PSNR, SSIM, and LPIPS fail to exhibit a monotonic trend. In contrast, DMD demonstrates a monotonic variation, effectively capturing the viewpoint discrepancy between views at a numerical level.


In our implementation, when computing $\text{DMD}(X^{src};X^{gen};X^{GT})$, the co-visibility threshold for calculating $P_{src}=\text{UFM}(X^{src}; X^{GT})$ is set to 0.3, while the threshold for $P_{gen}=  \text{UFM}(X^{gen}; X^{GT})$ is set to 0.2.

\begin{figure}[tb]
  \centering
  \includegraphics[width=1\linewidth]{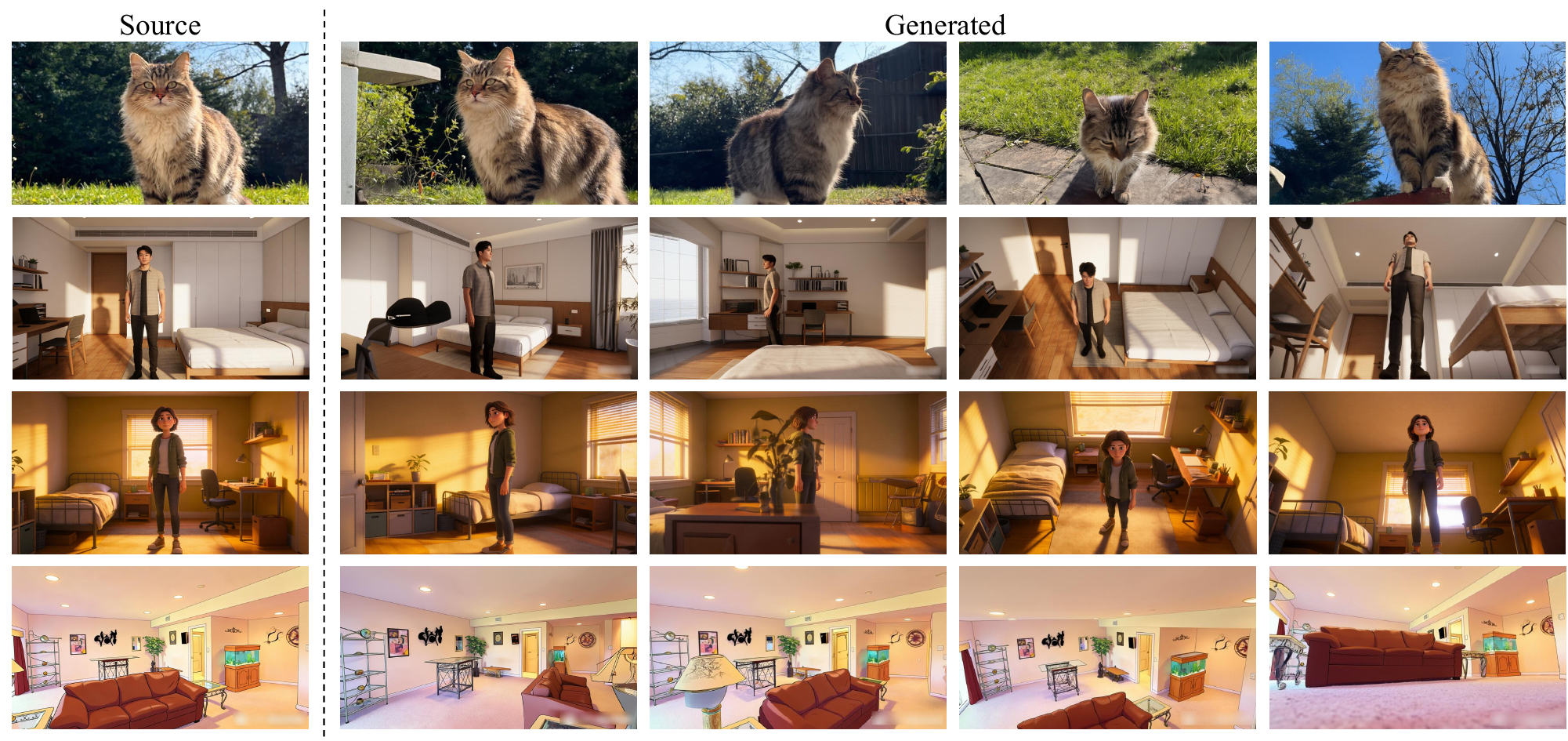}
  \caption{\textbf{Application on open-domain data.} Our method preserves the rich semantic knowledge of the pretrained model, enabling generalization across diverse scenes.}
  \label{fig:app}
      \vspace{-1.2em}
\end{figure}


\subsection{Application}
Our method also demonstrates strong generalization to diverse out-of-domain data. Since the pretrained backbone parameters remain frozen, the rich semantic priors acquired during large-scale pretraining are fully preserved. This design enables MetaView to maintain robust performance across a wide variety of scenes.
As shown in Fig. \ref{fig:app}, our approach produces consistent and accurate novel view synthesis results for scenes centered on humans and animals. Moreover, these examples span diverse visual styles, including cartoon, watercolor paintings, and AI-generated content. Such cross-domain robustness further highlights the practical applicability and downstream potential of MetaView.



\section{Conclusion}
In this paper, we present MetaView, a diffusion-based monocular novel view synthesis framework that achieves precise viewpoint control and large view generation without explicit 3D reconstruction. 
By bridging the gap between explicit geometric constraints and implicit generative modeling, MetaView effectively addresses the dual challenges of structural inconsistency and scale drifting. Specifically, we incorporate implicit geometry priors to guide fine-grained structural synthesis and preserve complex relative relationships, while introducing minimal yet essential 3D cues to anchor scene scale during generation. Extensive qualitative and quantitative experiments across multiple datasets demonstrate that our method consistently outperforms existing approaches. Moreover, MetaView exhibits strong generalization capability, including robustness to out-of-domain data. In future work, we will explore the spatial-aware generation in 4D scenes, enabling geometry-consistent in temporally evolving environments.

\section{Acknowledgments}
This research is supported by the MoE AcRF Tier 2 grant (MOE-T2EP20223-0001) and the MoE AcRF Tier 1 grant (RG14/22).

\clearpage  


%
%
\bibliographystyle{splncs04}
\bibliography{arxiv_main}

\clearpage 

\begin{center}
    \Large \textbf{Supplementary Material}
\end{center}
\vspace{1em}

\appendix
\setcounter{section}{0}
\renewcommand{\thesection}{\Alph{section}} 

\setcounter{figure}{0}
\renewcommand{\thefigure}{\Alph{figure}} 

\setcounter{table}{0}
\renewcommand{\thetable}{\Alph{table}} 

\setcounter{equation}{0}
\renewcommand{\theequation}{\Alph{equation}} 




\section{Implementation Details}
\subsection{Baseline Implementation}

\noindent \textbf{ViewCrafter}\cite{viewcrafter} We use the official open-source code of ViewCrafter\footnote{https://github.com/Drexubery/ViewCrafter}. All experiments are conducted under the single-image input setting, employing the official pre-trained model weights designed for generating 25-frame videos at a resolution of $1024 \times 576$. The camera poses and depth information required for its reconstruction module are obtained using VIPE\cite{vipe} estimation. Given the relative camera pose of the target viewpoint, we apply the pose interpolation method provided in the official codebase to generate a continuous 25-frame sequence of pose variations, which serves as the control condition. All other experimental settings follow the default configurations provided in the open-source code.

\noindent \textbf{Gen3C}\cite{gen3c} We use the official open-source code of Gen3C\footnote{https://github.com/nv-tlabs/GEN3C}. We employ the official pre-trained model weights to generate videos at a resolution of $1280 \times 720$, with a default sequence length of 121 frames. The camera poses and depth information required for its reconstruction module are obtained using VIPE\cite{vipe} estimation. We use the original camera trajectories from the test set as the control condition. To align with the required sequence length, if a trajectory contains fewer than 121 frames, it is padded to 121 frames by repeating the pose of the final frame. Conversely, if a trajectory exceeds 121 frames, it is uniformly sampled to exactly 121 frames. The input prompt is fixed to "A static scene". All other experimental settings follow the default configurations provided in the open-source code.

\noindent \textbf{Voyager}\cite{voyager} We use the official open-source code of Voyager\footnote{https://github.com/Tencent-Hunyuan/HunyuanWorld-Voyager}. We utilize the official open-source pre-trained model weights to generate 49-frame videos at a resolution of $768\times512$. The camera poses and depth information required for reconstruction are estimated using VIPE\cite{vipe}. Given the relative camera pose of the target viewpoint, we apply a uniform camera pose interpolation to obtain a continuous 49-frame sequence of pose variations. All other experimental settings follow the default configurations provided in the open-source code.

\noindent \textbf{PE-Field}\cite{PE-field} We use the official open-source code of PE-Field\footnote{https://github.com/MTLab/PE-Field}. We employ the official open-source pre-trained model for our experiments at a resolution of $960\times528$. The camera poses and depth information required for the warping process are estimated using VIPE\cite{vipe}. All other experimental settings follow the default configurations provided in the open-source code.


\noindent \textbf{HY-World}\cite{HY-world} We use the official open-source code of HY-World\footnote{https://github.com/Tencent-Hunyuan/HY-WorldPlay}. We employ the official open-source distilled model to generate 77-frame videos at a resolution of $832\times480$, using 4 inference steps. The camera poses are obtained from the VIPE\cite{vipe} estimation results on the test set. We apply a uniform camera pose interpolation to obtain a continuous sequence of pose variations corresponding to the 77 video frames. The input prompt is fixed to "A static scene". All other experimental settings follow the default configurations provided in the open-source code.

\noindent \textbf{Lingbot-World}\cite{lingbot-world} We use the official open-source code of Lingbot-World\footnote{https://github.com/robbyant/lingbot-world}. We employ the official open-source pre-trained camera control model to generate 41-frame videos at a resolution of $832\times480$. The camera poses are derived from the VIPE\cite{vipe} estimation results on the test set. A uniform camera pose interpolation is applied to obtain a continuous sequence of pose variations corresponding to 41 video frames. The input text prompt is fixed to "a static scene". All other experimental settings follow the default configurations provided in the open-source code.

To ensure a fair comparison and mitigate the sensitivity of low-level metrics to image resolution, all generated images are resized to $960\times528$ prior to evaluation.

\subsection{Data Curation}
We collect RGB data from the DL3DV\cite{dl3dv}, RealEstate10K\cite{realestate}, and Sekai-Real-Walk-HQ\cite{sekai} datasets. To obtain camera poses and depth information, we employ VIPE\cite{vipe}, utilizing DepthAnything3-Metric\cite{DA3} as a scale prior. This guarantees that the scale of the estimated depth is aligned with the corresponding camera poses. To ensure the quality of the training and evaluation data, we apply a rigorous filtering pipeline to remove low-quality or inconsistent scenes. All values below are in meters:
\begin{itemize}
    \item Invalid Depth: We first discard scenes containing anomalous depth values (e.g., NaN or Inf). These anomalies typically indicate estimation failures caused by shot transitions (cuts), overexposure, or underexposure.
    \item Expansive Distant Scenes: We exclude unbounded outdoor scenes, specifically defined as those where the minimum depth exceeds 50. The large numerical scale in such environments tends to compromise estimation accuracy and introduce excessive noise.
    \item Extreme Viewpoint Variations: We eliminate scenes where the maximum relative camera translation exceeds 100. Excessively large viewpoint shifts make it difficult for the method to maintain reliable pose estimation.
    \item Dynamic Scenes: The Sekai-Real-Walk-HQ dataset occasionally contains dynamic elements (e.g., pedestrians, animals, and moving vehicles) that disrupt the static scene assumption and geometric consistency. To address this, we leverage the vision-language model Qwen3-VL-32B-Instruct\cite{qwen3-vl} to automatically detect and filter out videos with dynamic objects. Specifically, we evaluate each video clip using the following prompt:

    {\sloppy \ttfamily "In this clip, are there any people, animals, moving vehicles, or video overlay subtitles that appear or disappear? Note: parked or stationary vehicles do not count; only vehicles in motion are considered. If any of the above are found, print 'False, [detected items]'; otherwise print 'True'."\par} Videos yielding a \texttt{"False"} output are subsequently discarded. This detection operation is applied iteratively to ensure comprehensive filtering of dynamic content.
\end{itemize}

\begin{figure}[tb]
  \centering
  \includegraphics[width=1\linewidth]{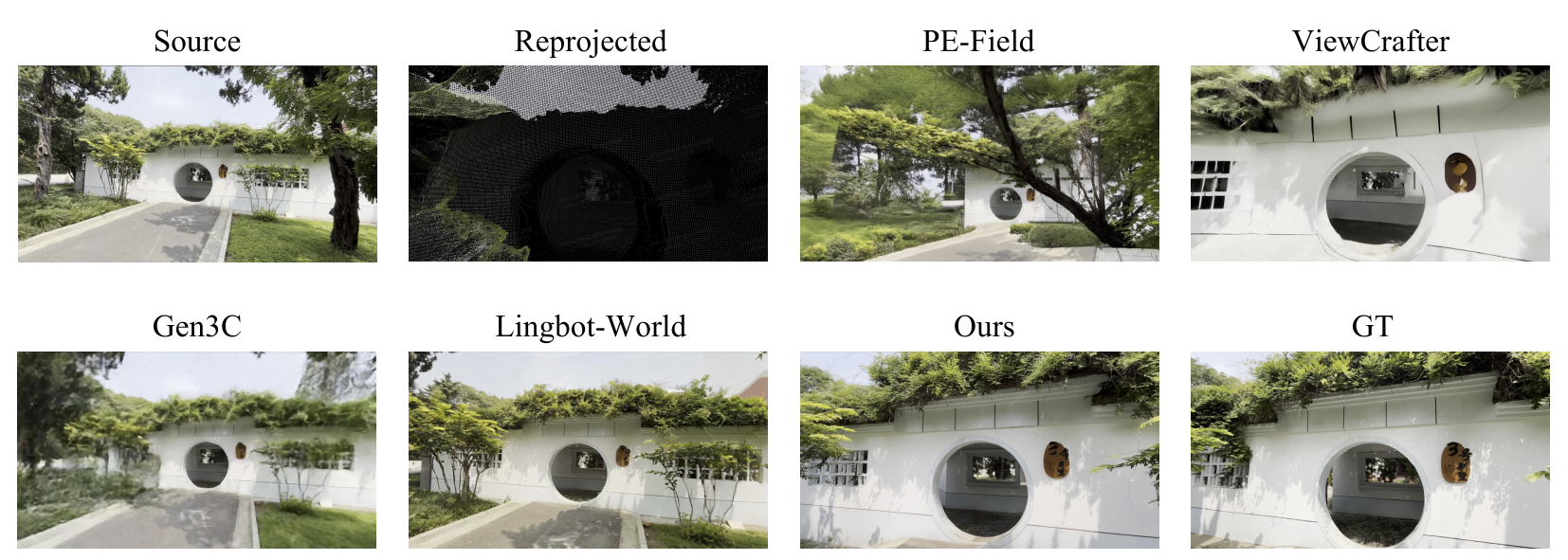}
  \caption{\textbf{Comparison of a zooming case.} Although the source image provides sufficient reference information, large viewpoint changes result in incomplete reprojection, causing certain reconstruction-based methods to fail in generating valid novel views.}
  \label{fig:suppl_view}
\end{figure}

\subsection{View Overlap Computation}
We argue that the difficulty of monocular NVS is determined by the amount of valid reference information the source view provides to the target view. Performing NVS when the source view offers no effective reference is practically ill-posed. Therefore, during both training and testing, we sample view pairs based on their view overlap ratio rather than relying on frame intervals to measure the extent of viewpoint variation. 

We employ a depth-based reprojection method to measure the view overlap ratio. Given the camera intrinsics $K$, target relative extrinsics $T$, and corresponding source and target depth maps $z^{src}, z^{tgt}$, we first unproject the pixels of the source view into 3D space. These 3D points are then transformed into the target image plane via $K$ and $T$ to obtain unprojected depth $\tilde{z}$. A source pixel is considered to be in the valid co-visible region if its corresponding 3D point satisfies three criteria: (1) it lies in front of the target camera ($\tilde{z}>0$), (2) its projected 2D coordinates fall strictly within the target image boundaries, and (3) it passes a depth consistency check. Specifically, the relative error between the $\tilde{z}$ and $z^{tgt}$ must be less than a threshold 10\%. The final overlap ratio is defined as the fraction of valid co-visible pixels over the total image resolution.
During training, we set a maximum frame interval of 40 and constrain the overlap ratio between the sampled source and target views to be at least 30\%. For evaluation, we partition the DL3DV test set into three difficulty levels—easy, medium, and hard— based on this overlap ratio. 

The view overlap ratio appropriately accounts for specific camera motions, such as continuous zooming. As shown in Fig. \ref{fig:suppl_view}, even with a large frame interval and substantial camera pose variation, the source view still provides the majority of the reference information. Consequently, the overlap ratio remains high, and these pairs are split into the easy subset. This characteristic also explains why baseline methods still exhibit sub-optimal performance even on the easy set, as the underlying large pose variations can disrupt their explicit geometric pipelines despite the high visual overlap.

\begin{figure}[tb]
  \centering
  \includegraphics[width=1\linewidth]{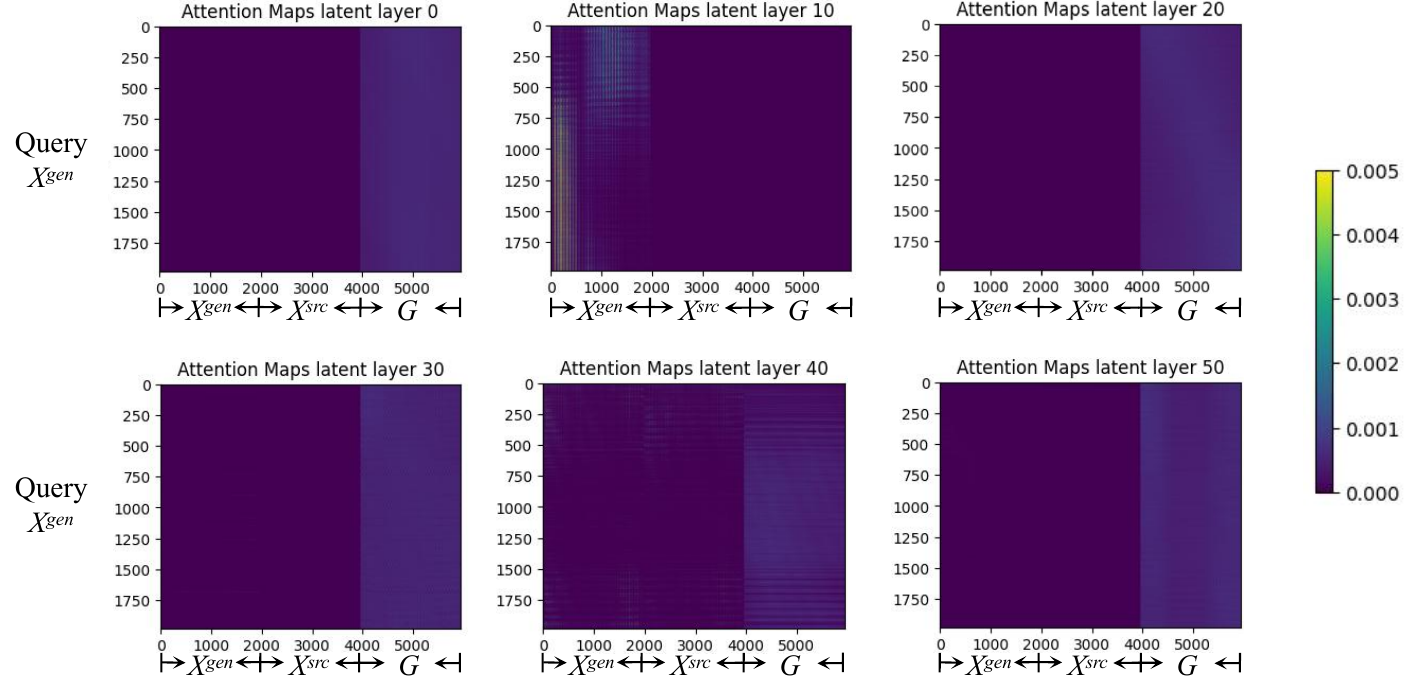}
  \caption{\textbf{Visualization of attention maps for $X^{gen}$.} The maps depict the attention weights averaged across all heads. $X^{gen}$ consistently allocates significantly higher attention (brighter regions) to the geometry token $G$ than to other components. }
  \label{fig:suppl_attn_map}
\end{figure}
\section{Effect of Geometry Tokens}
To further validate the effectiveness of the geometry token, we visualize the attention maps of the generation tokens $X^{gen}$ with respect to various components within the parallel self-attention layers. As illustrated in Fig. \ref{fig:suppl_attn_map}, which displays the attention weights averaged across all attention heads, $X^{gen}$ exhibits significantly higher attention (visibly brighter areas) towards the geometry token $G$ across all layers compared to other components. This observation indicates that during the synthesis process, $X^{gen}$ actively aggregates valid geometric reference information from $G$, demonstrating that $G$ plays a crucial guiding role in the generation.

\section{Limitations \& Discussion}
Constrained by the capabilities of the foundation model, MetaView may experience extrapolation failures in certain complex out-of-domain scenarios. This is primarily because the semantics of such scenes are absent from the pre-trained priors. Moreover, in expansive distant scenes where accurate scale estimation is inherently challenging, MetaView may suffer from imprecise viewpoint control. Furthermore, MetaView is currently restricted to viewpoint transformations within static environments; it struggles to generate valid and geometrically consistent results for 4D scenes containing dynamic objects. In future work, we plan to leverage video generation models to develop a spatio-temporally aware novel view synthesis framework for dynamic 4D scenes. Additionally, integrating the model's native text-conditioning mechanism into the synthesis pipeline to enable scene editing presents another promising direction for future exploration.

\section{More Results}
\begin{figure}
  \centering
  \includegraphics[width=0.8\linewidth]{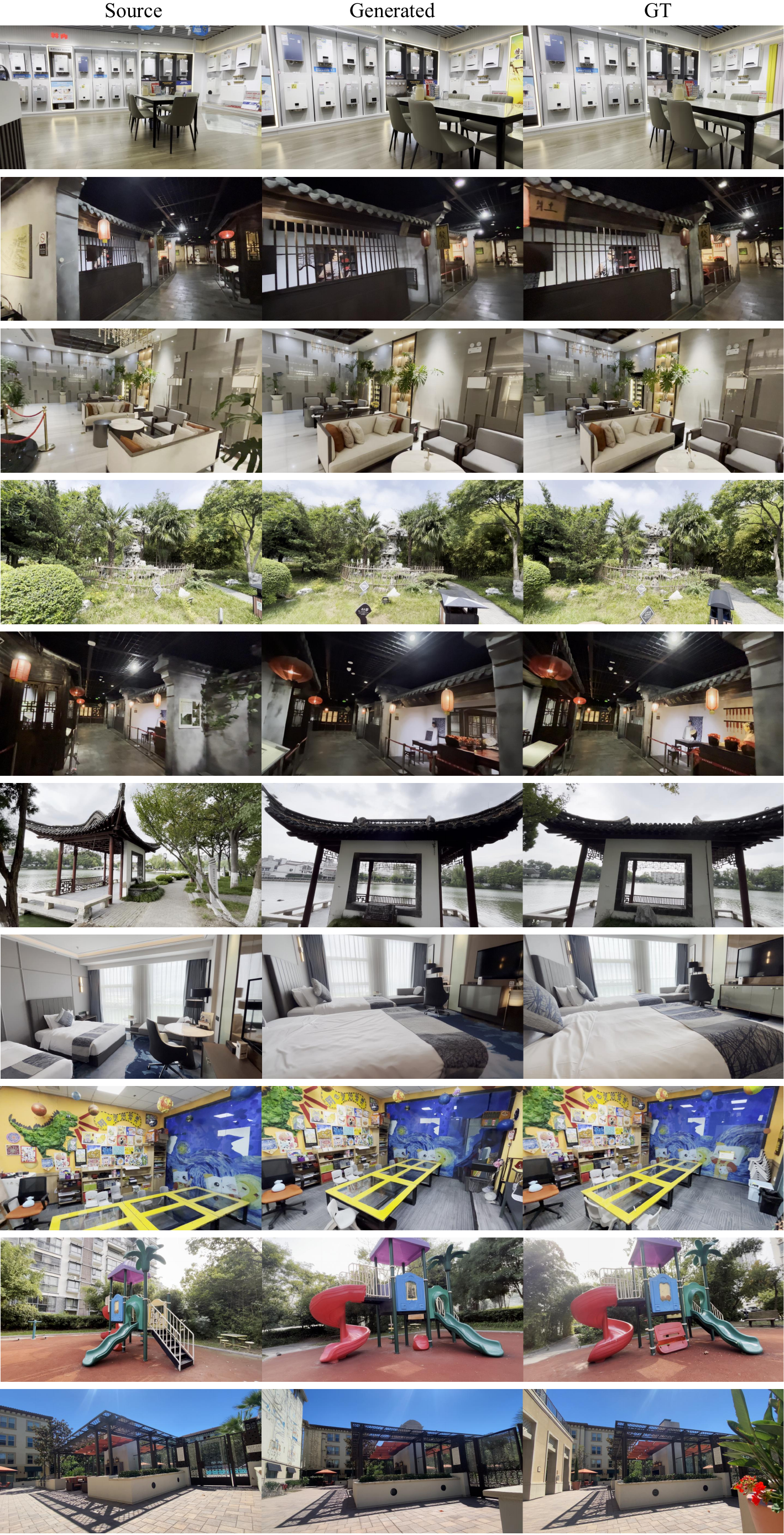}
  \caption{\textbf{Visualization results of MetaView on DL3DV.}}
  \label{fig:suppl_DL3DV}
\end{figure}

\begin{figure}
  \centering
  \includegraphics[width=0.8\linewidth]{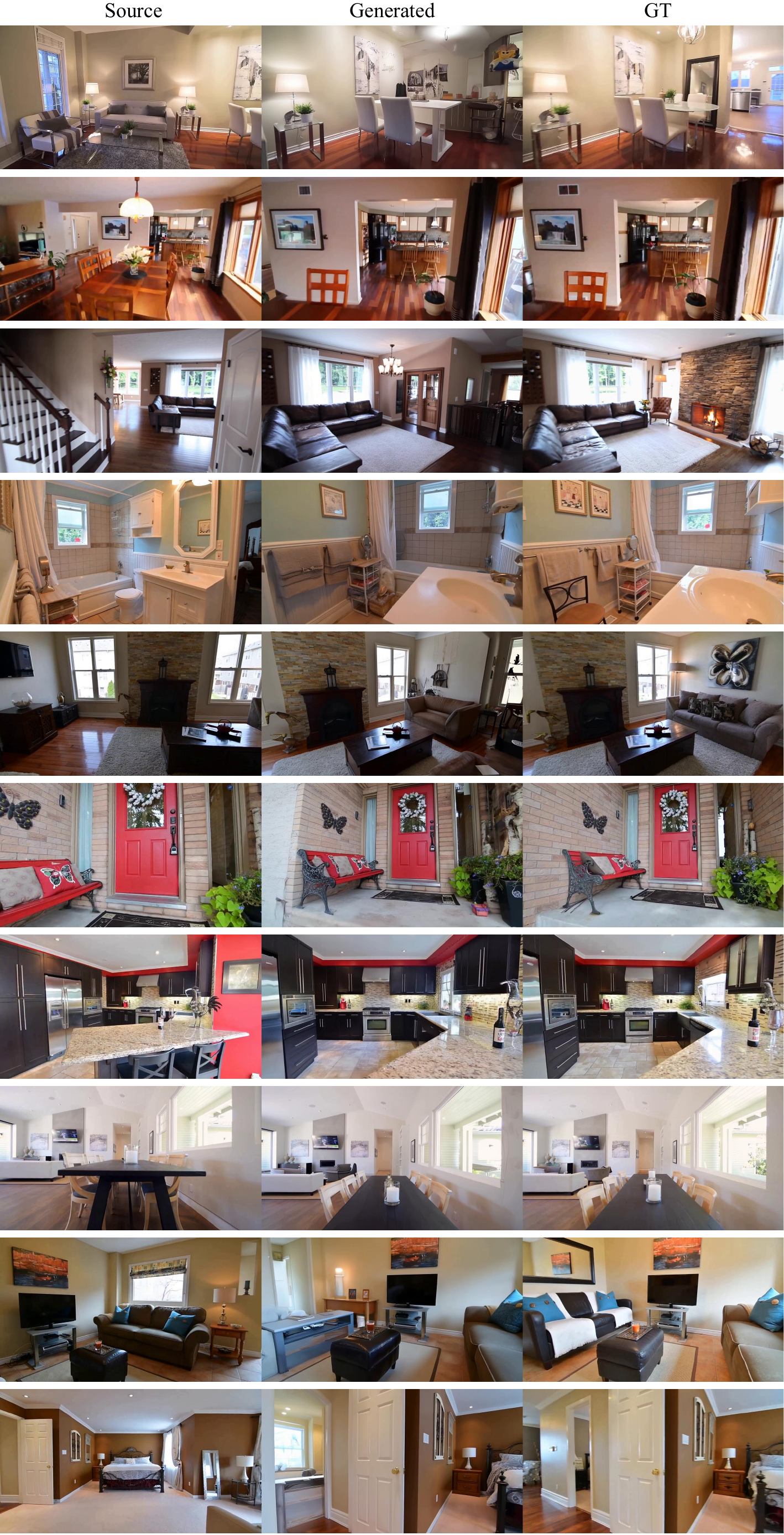}
  \caption{\textbf{Visualization results of MetaView on RealEstate10K.}}
  \label{fig:suppl_Re10K}
\end{figure}

We provide additional qualitative visualization results of MetaView. The results evaluated on the DL3DV\cite{dl3dv} (Fig. \ref{fig:suppl_DL3DV}), RealEstate10K\cite{realestate} (Fig. \ref{fig:suppl_Re10K}), and Sekai-Real-Walking-HQ\cite{sekai} (Fig. \ref{fig:suppl_Sekai}) test sets are illustrated in the following figures. Please zoom in for better visual assessment.

\begin{figure}
  \centering
  \includegraphics[width=0.8\linewidth]{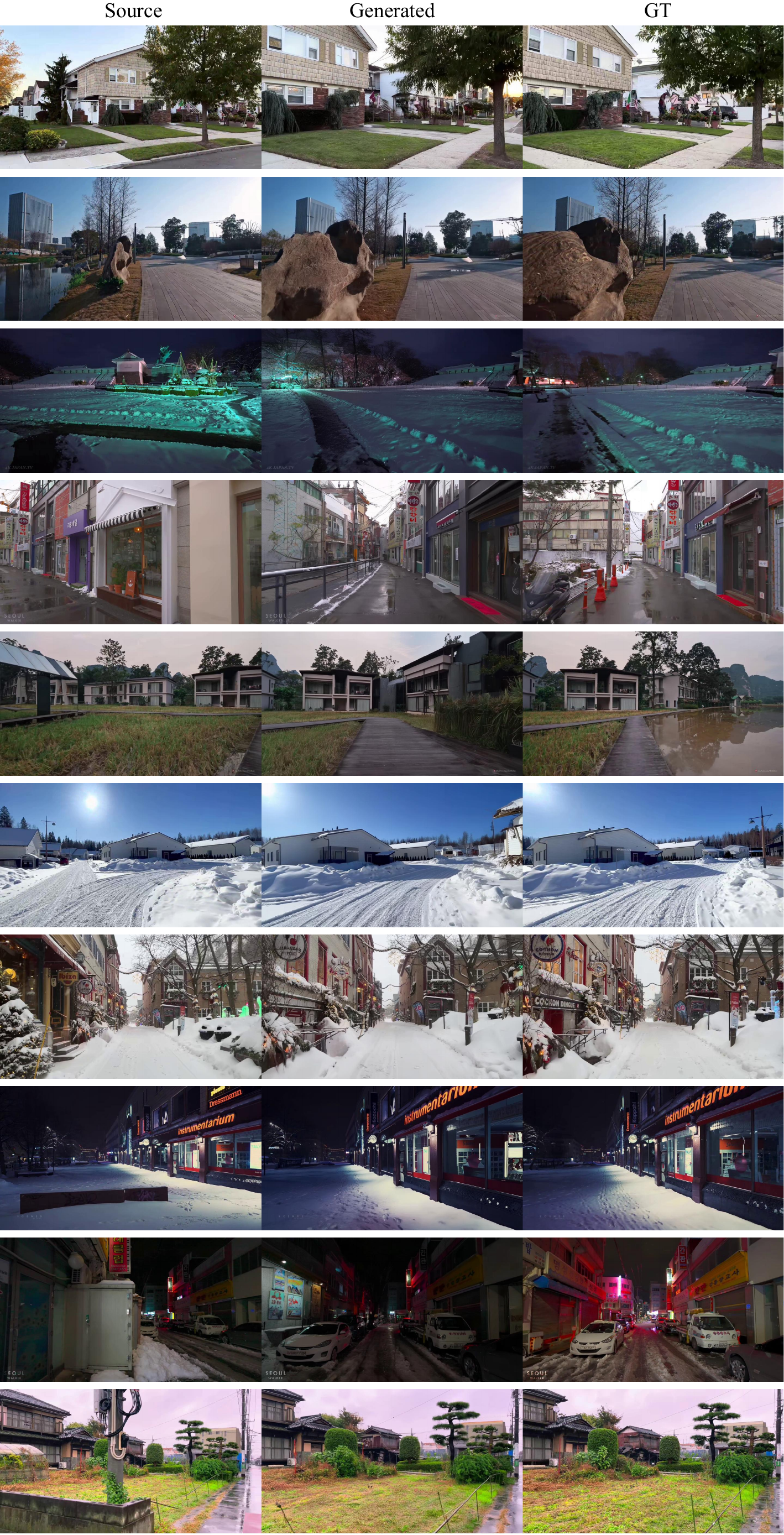}
  \caption{\textbf{Visualization results of MetaView on Sekai-Real-Walking-HQ.}}
  \label{fig:suppl_Sekai}
\end{figure}

\clearpage  

%
%

\end{document}